\documentclass{article}
\usepackage{ijcai19}

\usepackage{times}
\usepackage{soul}
\usepackage{url}
\usepackage[hidelinks]{hyperref}
\usepackage[utf8]{inputenc}
\usepackage[small]{caption}
\usepackage{graphicx}
\usepackage{amsmath}
\usepackage{subfigure} 
\usepackage{booktabs}
\usepackage{bbm}
\usepackage{amssymb}
\usepackage{algorithmic}
\usepackage{algorithm}
\usepackage{multirow}
\usepackage{array}
\def\Vec#1{{\boldsymbol{#1}}}
\def\Mat#1{{\boldsymbol{#1}}}

\title{Domain Adversarial Reinforcement Learning for Partial Domain Adaptation}

\author{
Jin Chen$^1$
\and
Xinxiao Wu$^1$\and
Lixin Duan$^{2}$\And
Shenghua Gao$^3$
\affiliations
$^1$Beijing Institute of Technology\\
$^2$University of Electronic Science and Technology\\
$^3$Shanghai University of Science and Technology
\emails
\{chen\_jin, wuxinxiao\}@bit.edu.cn,
lxduan@gmail.com,
gaoshh@shanghaitech.edu.cn
}
\begin{document}
\maketitle
\begin{abstract}
Partial domain adaptation aims to transfer knowledge from a label-rich source domain to a label-scarce target domain which relaxes the fully shared label space assumption across different domains. In this more general and practical scenario, a major challenge is how to select source instances in the shared classes across different domains for positive transfer. To address this issue, we propose a Domain Adversarial Reinforcement Learning (DARL) framework to automatically select source instances in the shared classes for circumventing negative transfer as well as to simultaneously learn transferable features between domains by reducing the domain shift. Specifically, in this framework, we employ deep Q-learning to learn policies for an agent to make selection decisions by approximating the action-value function. Moreover, domain adversarial learning is introduced to learn domain-invariant features for the selected source instances by the agent and the target instances, and also to determine rewards for the agent based on how relevant the selected source instances are to the target domain. Experiments on several benchmark datasets demonstrate that the superior performance of our DARL method over existing state of the arts for partial domain adaptation.
\end{abstract}

\section{Introduction}
\begin{figure*}[!t]
	\centering
	\includegraphics[width=1\textwidth]{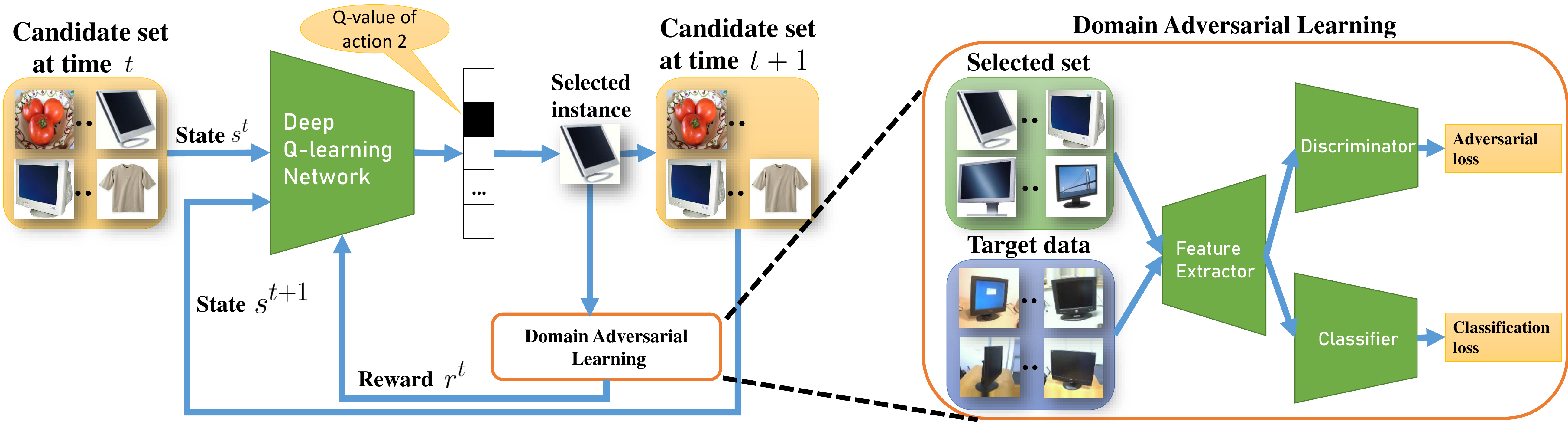}
	\caption{The framework of DARL. Deep Q-learning is used to learn policies for selecting source instances in the shared classes. A deep Q-learning network approximates the action-value function, and then the agent selects one instance according to the estimated Q-value. The reward is determined by how relevant the source instances are to the target domain measured by the domain adversarial learning.
	After $n$ selections, (\emph{i.e.} obtaining a selected set with $n$ source instances), both the selected set and the target data set are used to learn domain-invariant features via domain adversarial learning.}
	\label{fig:QDA-Nets}
\end{figure*}
Partial domain adaptation aims at leveraging the label-rich domain (source domain) to boost the performance of the label-scarce domain (target domain), where the target label space is a subspace of the source label space.
Partial domain adaptation relaxes the fully shared label space assumption in domain adaptation, which makes it more general and practical with growing attention.

Directly matching the feature distributions between the source and target domains~\cite{Long2015Learning,ganin2016domain,Shen2018Wasserstein} for partial domain adaptation  will lead to negative transfer due to the mismatch of label spaces across different domains. 
To solve this problem, existing methods of partial domain adaptation resort to up-weighting source instances in the shared classes while down-weighting source instances in the outlier classes (\emph{i.e.,} classes that the target domain does not contain).~\cite{zhang2018importance} applied a two-domain classifier  to identify the weights of source instances. In~\cite{Cao_2018_CVPR,Cao_2018_ECCV}, the weights of source instances are determined by the class probability distribution of target instances based on their prediction scores obtained from the source classifier.

In this paper, we propose to select source instances from the shared classes and use them as anchors to learn an adaptive classifier for the target domain. Since there is no labels available in the target domain for the partial domain adaptation task, it is nontrivial for us to perform the source instance selection. Unlike some existing works which select instances based on their pseudo labels~\cite{Cao_2018_ECCV,Cao_2018_CVPR,zhang2018importance}, we employ the reinforcement learning paradigm to automate the instance selection procedure.

Specifically, we propose a Domain Adversarial Reinforcement Learning (DARL) framework for partial domain adaptation, which couples deep Q-learning with domain adversarial learning~\cite{ganin2016domain}.
The deep Q-learning learns policies for selecting source instances in the shared classes with the relevance of source instances to the target domain as rewards.
The domain adversarial learning is introduced to learn domain-invariant features for the selected source instances and the target instances, and simultaneously to determine rewards for guiding the selection of the agent based on the relevance.
Concretely, a deep Q-learning network is built to approximate the action-value function, which takes state as input and outputs Q-values of different actions. Actions are corresponding to source instances, and states are represented by feature vectors of those source instances. 
According to the Q-values estimated by the deep Q-learning network, the agent takes one action and the reward of this action is provided by a domain adversarial learning network. 
After several selections, the selected source instances and the target instances are used for updating the domain adversarial learning network to learn domain-invariant features, where a discriminator aims to distinguish the source domain from the target domain, and a feature extractor tries to confuse the discriminator to make the features as indistinguishable as possible.
An iterative optimization algorithm is proposed to jointly train the deep Q-learning network and the domain adversarial learning network in an end-to-end manner.

The main contributions are summarized as follows:
\begin{itemize}
\item{We propose a new framework called Domain Adversarial Reinforcement Learning (DARL) for partial domain adaptation. With the superior exploration ability of reinforcement learning and the good performance of domain adversarial learning on the domain shift reduction, DARL is able to automatically select source instances in the shared classes and simultaneously learn transferable features between different domains.}
\item{We design a novel reward  based on domain adversarial learning in DARL, which guides the agent to learn right selection policies by measuring the relevance of the source instances to the target domain.}
\item{Evaluations on various benchmark datasets demonstrate that DARL achieves superior results than existing state-of-the-arts for partial domain adaptation.}
\end{itemize}

\section{Related Work}
Existing partial domain adaptation methods focus on up-weighting source instances in the shared classes or improving the importance of shared source classes.
\cite{Cao_2018_CVPR} introduced multiple discriminators for fine-grained adaptation, where the class probability of each instances modeled by the source classifier is used as the weights for domain discriminators. 
By multiple probability weighted domain discriminators, each instance is aligned with relevant classes. 
\cite{Cao_2018_ECCV} extended the domain adversarial network with the weight of each source class which is computed with the class probability of target data predicted by the source classifier.
\cite{zhang2018importance} introduced a two domain classifier framework, where the weights of source instances are given by the domain scores predicted by the first domain classifier and the second domain classifier is applied to reduce the domain shift between weighted source instances and the target instances.
Different from those methods, we propose domain adversarial reinforcement learning to select source instances in the shared classes. With the domain adversarial learning based reward, the agent can automatically learn the selection policies by reducing the domain shift between the source and target domains.

Reinforcement Learning~\cite{Sutton1998Reinforcement} has made great process in many vision tasks, such as video caption~\cite{Wang_2018_CVPR},
action recognition~\cite{Yeung_2017_CVPR}, object tracking~\cite{Yun2017Action,Ren_2018_ECCV_Collaborative} and detection~\cite{Huang2017A,Pirinen_2018_CVPR}. 
\cite{Yeung_2017_CVPR} applied the reinforcement learning for action recognition with the wild videos, where the agent aims to select videos similar with seed videos from noisy web search results.
\cite{dong2018domain} introduced a policy network for selecting source images similar with a random target instance for one shot learning.
Different from the aforementioned works, we apply reinforcement learning to partial domain adaptation for selecting source instances in the shared classes. A novel reward based on domain adversarial learning is proposed to provide effective guidance to the agent.

\section{Domain Adversarial Reinforcement Learning}
For the partial domain adaptation in unsupervised scenario, we are given a labeled source domain $\mathcal{D}_s=\{(x_{i}^s,y_{i}^s)|_{i=1}^{N_s}\}$ drawn i.i.d from the source distribution $p(x)$ with $y_{i}^s \in \mathcal{Y}_s$ and an unlabeled target domain $\mathcal{D}_t = \{x_{j}^t|_{j=1}^{N_t}\}$ drawn i.i.d from the target distribution $q(x)$. $N_s$ and $N_t$ are the numbers of instances in the source and target domains, respectively.
The target class label space $\mathcal{Y}_t$ is a subspace of the source class label space $\mathcal{Y}_s$, \emph{i.e.,} $\mathcal{Y}_t \subset \mathcal{Y}_s$. 
The classes in $\mathcal{Y}_s$ but not in $\mathcal{Y}_t$ are denoted as outlier classes, and the common classes in $\mathcal{Y}_s$ and $\mathcal{Y}_t$ are denoted as shared classes.
The data distributions of source and target domains are different, \emph{i.e.,} $p(x) \ne q(x)$. 
The Domain Adversarial Reinforcement Learning (DARL) framework is proposed to select source instances with the class labels $y_{i}^s \in \mathcal{Y}_t$ and learn transferable features of the selected source instances and target instances in the shared label space $\mathcal{Y}_t$. The architecture of DARL is shown in Figure \ref{fig:QDA-Nets}.

\subsection{Deep Q-learning}
The deep Q-learning is applied to learn policies for selecting source instances in the shared classes.
We define a candidate set $\mathcal{D}_c$ which consists of source instances to be selected and is initialized as the randomly sampled instances from the source domain, and a selected set $\mathcal{D}_e$ which is constructed by the selected source instances and initialized to empty.
At timestep $t$, the agent takes an action $a_t$ according to the Q-value $Q(\Mat{s}_t,a)$ estimated by the deep Q-learning network with the state $\Mat{s}_t$ as input.
The action $a_t$ is equivalent to selecting the corresponding instance from the candidate set $\mathcal{D}_c$ and moving it to the selected set $\mathcal{D}_e$.
The reward $R_t$ of action $a_t$ and the next state $s_{t+1}$ are sent to the agent for the next selection. 
This is one selection process of the agent.
In each episode of deep Q-learning, the agent makes several selections until it reaches the terminal state on the candidate set.

\noindent\textbf{State.} At the initial of one episode, given the candidate set $\mathcal{D}_c=\{(x_{i}^c,y_{i}^c)|_{i=1}^{N_c}\}$ with $N_c$ instances and the initial selected set $\mathcal{D}_e = \varnothing$, the initial state $\Mat{s}_0$ is constructed by the feature vectors of instances in $\mathcal{D}_c$, represented by $\Mat{s}_0 = [F(x_{1}^c),\cdots,F(x_{N_c}^{c})]\in \mathbb {R}^{d \times N_c}$, where $F(x_i^c)$ denotes the $d$-dimensional feature vector of instance $x_i^c$ extracted by the feature extractor $F$ of the domain adversarial learning network.
After taking an action, the corresponding instance in $\mathcal{D}_c$ is moved from $\mathcal{D}_c$ to $\mathcal{D}_e$. Thus, the size of state is changed from $d\times N_c$ to $d \times (N_c-1)$. 
In order to keep the size of state constant, we replace the selected instance with a zero-valued feature vector.

\noindent\textbf{Action.} The action is defined as selecting one instance from the candidate set $\mathcal{D}_c$. At each timestep, the agent takes one action from the action set $A = \{a_1,a_2,\ldots,a_{N_c}\}$, where $a_i$ means that selecting the $i$-th instance in $\mathcal{D}_c$ and then moving it to $\mathcal{D}_e$. 
The number of actions is the same as the number of instances in $\mathcal{D}_c$, \emph{i.e., $N_c$}. 
The optimal action taken by the agent at timestep $t$ is formulated by 
\begin{equation}
\label{equation:policy}
\begin{aligned}
a_t = \max \limits_{a} Q(\Mat{s}_t,a),
\end{aligned}
\end{equation}
where $s_t$ indicates the state at timestep $t$ and the Q-value $Q(s_t,a)$ is the accumulated rewards of taking the action $a$.
A deep Q-learning network is introduced to estimate $Q(s_t,a)$.
It uses $s_t$ as input and outputs a $|A|$-dimensional vector which represents the Q-values of $|A|$ actions.

\noindent\textbf{Reward.}
The reward is the feedback of the corresponding action taken by the agent. It guides the agent to make selection decisions. 
Since the source instances in the shared classes should be more relevant to the target domain than the source instances in the outlier classes, we use the relevance of source instances to the target domain to design the reward.

When the agent takes the action $a_t$ to move the candidate instance $x$ to the selected set $\mathcal{D}_e$, the reward of the action $a_t$ is computed by
\begin{equation}
\label{equation:reward}
\begin{aligned}
&R_t=
\begin{cases}
+1, \mbox{if       }  \varphi(x)> \tau\\
-1, \mbox{otherwise} 
\end{cases}
\end{aligned}
\end{equation}
where $\varphi(x)$ is a metric function of measuring the relevance of instance $x$ to the target domain, and will be detailed in Section \ref{section:relevance-metric}. The more relevant the instance $x$ is to the target domain, the higher the value of $\varphi(x)$ becomes. We adopt a binary reward, \emph{i.e.,} $+1$ and $-1$, which has been widely used in reinforcement learning for various tasks \cite{Yun2017Action,Ren_2018_ECCV} . Because a binary reward can help the agent clearly distinguish good or bad actions and provide more explicit guidance than directly using the relevance measure as a reward. If directly using the relevance measure, the relevance difference between different instances is too small to confuse the agent about which actions are good and which actions are bad. If $\varphi(x)$ is higher than the threshold $\tau$, then the reward for the agent will be $+1$, otherwise the reward will be $-1$. When the reward is $-1$, the agent reaches the terminal state, stops the selection on the current candidate set, and begins a new selection on the next candidate set.

\noindent\textbf{Objective function.}
Based on the definitions of the state, action and reward, the objective function of deep Q-learning network is given by
\begin{equation}
\label{equation:DQN_loss}
\begin{aligned}
\mathcal{L}_{q} =\mathbb{E}_{\Mat{s}_t,a_t}\bigg[
\Big(V(\Mat{s}_t)-Q\big(\Mat{s}_t,a_t\big)\Big)^2 \bigg],
\end{aligned}
\end{equation}
where $V(\Mat{s}_t)-Q(\Mat{s}_t,a_t)$ is the temporal difference error.  $V(\Mat{s}_t)$ is the target value of $Q(\Mat{s}_t,a_t)$, estimated by
\begin{equation}
\label{equation:nextstate}
\begin{aligned}
V(\Mat{s}_t)=\mathbb{E}_{\Mat{s}_{t+1}}\Big[
R_t + \gamma \max_{a_{t+1}} Q\big(\Mat{s}_{t+1},a_{t+1}|\Mat{s}_t,a_t\big)\Big],
\end{aligned}
\end{equation}
where the first term $R_t$ is the reward of taking the action $a_t$, computed by Eq.\eqref{equation:reward}, and the second term is the future reward estimated by the current deep Q-learning network with the next state $\Mat{s}_{t+1}$.

\subsection{Domain Adversarial Learning}
\label{section:DA}
The goal of domain adversarial learning is to learn transferable features for reducing the domain shift, which is achieved by the adversarial learning procedure of a discriminator $D$ and a feature extractor $F$. 
The discriminator $D$ is trained to distinguish the source domain from the target domain, and the feature extractor $F$ is trained to confuse the discriminator $D$.
Thus, the adversarial loss of domain adversarial learning is summarized as the minimax form:
\begin{equation}
\label{equation:L_d}
\begin{aligned}
\min \limits_{F} \max \limits_{D} \mathcal{L}_d(F,D)&= \mathbb{E}_{(x)\sim p(x)}\log\bigg(D\big(F(x)\big)\bigg)\\
&+\mathbb{E}_{(x)\sim q(x)}\mathrm{log}\bigg(1-D\big(F(x)\big)\bigg).
\end{aligned}
\end{equation}
With the fixed $F$, the discriminator $D$ learns an optimal bound of the true domain distribution by maximizing the adversarial loss $\mathcal{L}_d(F,D)$.
With the optimal discriminator $D$, the feature extractor $F$ is trained for more domain-invariant feature by minimizing the adversarial loss $\mathcal{L}_d(F,D)$.
With the transferable features, an adaptive classifier $C$ is trained by minimizing the following source risk: 
\begin{equation}
\label{equation:L_c}
\begin{aligned}
\min \limits_{F,C} \mathcal{L}_c(F,C)
= \mathbb{E}_{x\sim p(x)}\left[- \sum_{k=1}^{K} \mathbbm{1}_{k=y} \log C(F(x))\right],
\end{aligned}
\end{equation}
where $y$ is the class label of instance $x$, and $K$ is the number of source classes, \emph{i.e.,} $K=|\mathcal{Y}_s|$. $\mathbbm{1}_{k=y}$ means that if $k=y$, the value of $\mathbbm{1}_{k=y}$ is 1 and otherwise is 0.

An optimal discriminator does not only distinguish the source domain from the target domain, but also identifies the category of source instances. 
To this end, a $K+1$-way classifier $D$ is introduced as the discriminator. The first $K$ ways model the class distribution, and the last way models the domain distribution. 
We use one-hot encoding to represent the category label of each instance $x$ and add an element to represent whether $x$ is from the target domain or not. 

When optimizing $D$ with the fixed $F$, the objective function of discriminator $D$ is
\begin{equation}
\label{equation:D_LOSS}
\begin{aligned}
\min \limits_{D} \mathcal{L}_{y}(F,D)=\mathbb{E}_{x\sim p(x)}H\bigg(D\big(F(x)\big),\Vec{\tilde{y}}_d^s\bigg)\\
+\mathbb{E}_{(x)\sim q(x)}H\bigg(D\big(F(x)\big),\Vec{\tilde{y}}_d^t\bigg),\\
\end{aligned}
\end{equation}
where $H(\cdot,\cdot)$ is the cross entropy loss. The source instance label $\Vec{\tilde{y}}_d^s$ and the target instance label $\Vec{\tilde{y}}_d^t$ are 
\begin{equation}
\label{equation:d-label}
\begin{aligned}
&\Vec{\tilde{y}}_d^s = [\overbrace{\underbrace{0,\cdots,0,1}_i,0,\cdots,0,0}^K,0], (x,y) \in \ \mathcal{D}_s,y = i,\\
&\Vec{\tilde{y}}_d^t = [\overbrace{0,\cdots\cdots\cdot\cdot\cdots\cdot\cdots\cdot,0}^K,1], x \in \mathcal{D}_t,\\
\end{aligned}
\end{equation}
where $y$ is the class label of source instance $x$. 

When optimizing $F$ with fixed $D$, the objective function of feature extractor $F$ is
\begin{equation}
\label{equation:F_LOSS}
\begin{aligned}
\min \limits_{F} \mathcal{L}_{y}(F,D)=\mathbb{E}_{x\sim p(x)}H\bigg(D\big(F(x)\big),\Vec{\tilde{y}}_f^s\bigg)\\
+\mathbb{E}_{(x)\sim q(x)}H\bigg(D\big(F(x)\big),\Vec{\tilde{y}}_f^t\bigg).
\end{aligned}
\end{equation}
The source instance label $\Vec{\tilde{y}}_f^s$ and the target instance label $\Vec{\tilde{y}}_f^t$ are 
\begin{equation}
\label{equation:f-label}
\begin{aligned}
&\Vec{\tilde{y}}_f^s = [\overbrace{0,\cdots\cdots\cdot\cdot\cdots\cdot\cdots\cdot,0}^K,1], (x,y) \in \mathcal{D}_s,\\
&\Vec{\tilde{y}}_f^t = [\overbrace{\underbrace{0,\cdots,0,1}_j,0,\cdots,0,0}^K,0], x \in \mathcal{D}_t,\hat{y} = j,
\end{aligned}
\end{equation}
where $\hat{y}$ is the pseudo label of target instance $x$ predicted by the classifier $C$.

We expect that the discriminator $D$ can classify labeled source instances and assign unlabeled target instances into the target domain. Thus, in Eq.\eqref{equation:d-label}, $\Vec{\tilde{y}}_d^s$ contains the source class information while $\Vec{\tilde{y}}_d^t$ does not when optimizing $D$ by Eq.\eqref{equation:L_d}. The feature extractor $F$ aims to confuse $D$, i.e., makes $D$ classify target instances into $K$ source classes and assign source instances into the target domain. Thus, in Eq.\eqref{equation:f-label}, $\Vec{\tilde{y}}_f^t$ contains the target class information while $\Vec{\tilde{y}}_f^s$ does not when optimizing $F$ by Eq.\eqref{equation:F_LOSS}. Similar manners are used in \cite{hu2018duplex,sankaranarayanan2018generate}.

\noindent\textbf{Objective function.}
The overall optimization problem of domain adversarial learning is as follows:
\begin{equation}
\label{equation:DAN}
\min \limits_{F,C,D} \,\, \mathcal{L}(F, C, D) =  \mathcal{L}_c(F,C)+\mathcal{L}_{y}(F,D).
\end{equation}
The feature extractor $F$ and the discriminator $D$ are trained in an adversarial manner with different label values of instances by minimizing the adversarial loss $\mathcal{L}_{y}(F,D)$.

\subsection{Relevance Metric}
\label{section:relevance-metric}
The relevance metric function $\varphi(x)$ measures the relevance of input instance $x$ to the target domain, which is based on the discriminator $D$ and the classifier $C$.

\noindent\textbf{Instance-level relevance measured by $D$.}
If the source instance is likely to be assigned into the target domain by the discriminator $D$, the relevance of this instance to the target domain is high.
The last element of the output of $D$ is denoted as $D(\cdot)_{d}$. 
The higher the $D(F(x))_{d}$ is, the more relevant the source instance $x$ is to the target domain.

\noindent\textbf{Class-level relevance measured by $C$.}
Since the target classes and the outlier classes have no overlap, the target data has low probability to be assigned into the outlier classes. 
Therefore, we use the predicted class distribution of the target data to compute the relevance of source classes to the target domain, denoted as
$\Vec{\mu}=[\mu_1,\mu_2,\ldots,\mu_K]\in \mathbb{R}^K$, where $\mu_i$ represents the relevance of the $i$-th source class to the target domain. The higher the $\mu_i$ is, the more relevant the $i$-th source class is to the target domain. 
We compute $\Vec{\mu}$ by
\begin{equation}
\label{equation:rc}
\begin{aligned}
\Vec{\mu} =\frac{1}{N_t} \sum_{i=0}^{N_t}C(F(x_i)), x_i \in \mathcal{D}_t,
\end{aligned}
\end{equation}
and normalize it by $\Vec{\mu} = \frac{\Vec{\mu}}{\rm{max}(\Vec{\mu})}$.

The instance-level and the class-level relevance represent the relevance of the instance $x$ to the target domain from different aspects. The bigger the values of the two terms are, the more relevant x is to the target domain. Thus, it is a natural way to compute the product of the two terms to evaluate of the relevance of $x$ to the target domain. The relevance metric function $\varphi(x)$ is given by
\begin{equation}
\label{equation:relevance function}
\begin{aligned}
\varphi(x) = \mu_i D\big(F(x)\big)_{d}, 
\end{aligned}
\end{equation}
where $i$ is the class label of source instance $x$.
Algorithm ~\ref{algorithm:RAN} summarizes the detailed algorithm of DARL.
\vspace{-2.5mm}
\begin{algorithm}
	\caption{DARL}
	\label{algorithm:RAN}
	\begin{algorithmic}[1]
		\renewcommand{\algorithmicrequire}{\textbf{Input:}}
		\REQUIRE Source domain $\mathcal{D}_s$ and target domain $\mathcal{D}_t$
		\renewcommand{\algorithmicensure}{\textbf{Output:}}
		\ENSURE The optimal $F$, $C$.
		\STATE Pre-train $F$ and $C$ with $\mathcal{D}_s$ by Eq.\eqref{equation:L_c};
		\STATE Initialize the experience pool $M=\varnothing$;
		\WHILE {not converge}
		\STATE Initialize $\mathcal{D}_c$, $\mathcal{D}_e$ and generate the state $\Mat{s}_0$ with $\mathcal{D}_c$;
		\WHILE {$\mathcal{D}_c \ne \varnothing$}
		\STATE Take an action $a_t$ using the policy Eq.\eqref{equation:policy};
		\STATE Compute the reward $R_t$ of $a_t$ by Eq.\eqref{equation:reward};
		\STATE Update $\mathcal{D}_c$, $\mathcal{D}_e$ and state;
		\STATE Insert recording $(\Mat{s}_t,a_t,\Mat{s}_{t+1},R_t)$ into $M$;
		\STATE Sample recordings from $M$ to update deep Q-learning network by Eq.\eqref{equation:DQN_loss};
		\STATE If $R_t<0$: break;
		\ENDWHILE
		\STATE Update $C,F,D$ with $\mathcal{D}_e$ and $\mathcal{D}_t$ by Eq.\eqref{equation:DAN}.
		\ENDWHILE
	\end{algorithmic}
\end{algorithm}

\begin{table*}
	\centering
    \scriptsize
	\begin{tabular}{|l|c|c|c|c|c|c|c|c|c|c|c|c|c|}
		\hline
		\multirow{3}{*}{Method} & \multicolumn{13}{c|}{Office+Caltech-10} \\
		\cline{2-14}
		& C10 $\to$  & C10 $\to$  & C10 $\to$ & A10 $\to$ &  A10 $\to$  & A10 $\to$  & W10 $\to$ & W10 $\to$  & W10 $\to$  &D10 $\to$  &D10 $\to$ &D10 $\to$  &\multirow{2}{*}{Avg.}\\
		
		& A5&W5& D5& C5&W5&D5& C5&A5&D5 &C5&A5&W5&\\
		\hline
		AlexNet+bottleneck &94.65&90.37&97.06&85.79&81.48&95.59&76.37&87.79& \textbf{100.00}& 80.99&89.94&97.04&89.76\\
		DANN~\cite{ganin2016domain} & 91.86&82.22&83.82&77.57& 65.93&80.88&72.60&80.30&95.59&69.35&77.09&80.74&79.83\\
		RTN~\cite{Long2016Unsupervised} &91.86& 93.33&80.88&80.99&69.63& 70.59&59.08 &74.73&\textbf{100.00}&59.08&70.02&91.11&78.44\\
		ADDA~\cite{Tzeng_2017_CVPR} &93.15&94.07&97.06&85.27&87.41&89.71&86.82&92.08&\textbf{100.00}&89.90&93.79&98.52&92.31\\
		IWAN~\cite{zhang2018importance} &94.22&97.78&98.53&89.90&87.41&88.24&90.24&95.29&\textbf{100.00}&91.61&94.43&98.52&93.85\\
		\hline
		DARL w/o Q-learning &94.86&97.04&\textbf{100.00}&86.13&88.15&97.06&85.96&93.36&\textbf{100.00}&85.79&89.08&99.26&93.06\\
		DAL with pseudo labels &95.29&91.85&98.53&90.24&85.93&95.59&91.27&95.29&
		\textbf{100.00}&80.99&89.94&98.52&92.79\\
		DARL &\textbf{96.36}&\textbf{98.52}&\textbf{100.00}&\textbf{92.47}&\textbf{88.89}&\textbf{100.00}&\textbf{93.15}&\textbf{96.15}&\textbf{100.00}&\textbf{92.64}&\textbf{95.93}&\textbf{99.26}&\textbf{96.11}\\
		\hline
	\end{tabular}
	\vspace{-2.5mm}
    \caption{Classification accuracies ($\%$) of partial domain adaptation tasks on Office+Caltech-10 (AlexNet as base network).}
	\label{table:results on Office+Caltech-10}
\end{table*}
\begin{table*}
	\centering
	\scriptsize
	\begin{tabular}{|c|l|c|c|c|c|c|c|c|}
		\hline
		\multirow{2}*{Base net}&\multirow{2}*{Method}& \multicolumn{7}{c|}{Office-31} \\
		\cline{3-9}
		&& A31 $\to$ W10 & D31$\to$  W10& W31 $\to$ D10 & A31 $\to$ D10&  D31 $\to$ A10 & W31 $\to$A10 &Avg\\
		\hline
		\multirow{10}*{AlexNet}&AlexNet+bottleneck &59.32&96.27&98.73&73.25&70.77&66.08&77.40\\
		&DAN~\cite{Long2015Learning} &56.52&71.86&86.78&51.86&50.42&52.29&61.62\\
		&DANN~\cite{ganin2016domain} & 56.95&75.59&89.17&57.32&57.62&63.15&66.64\\
		&RTN~\cite{Long2016Unsupervised} &66.78&86.77&99.36&70.06&73.52&76.41&78.82\\
		&ADDA~\cite{Tzeng_2017_CVPR} &70.68&96.44&98.65&72.90&74.26&75.56&81.42\\
		&IWAN~\cite{zhang2018importance} &76.27&98.98&\textbf{100.00}&78.98&89.46&81.73&87.57\\
		&SAN~\cite{Cao_2018_CVPR} &\textbf{80.02}&98.64&\textbf{100.00}&81.28&80.58&83.09&87.27\\
		\cline{2-9}
		&DARL w/o Q-learning &60.00&97.63&98.09&75.08&81.52&78.50&81.92\\
		&DAL with pseudo labels &67.46&98.89&99.36&73.98&90.71&81.94&85.39 \\
		&DARL &77.97&\textbf{100.00}&\textbf{100.00}&\textbf{82.80}&\textbf{93.01}&\textbf{87.47}&\textbf{90.21}\\
		\hline
		\multirow{9}*{ResNet-50}&ResNet+bottleneck &74.58&94.58&95.54&78.34&70.77&70.56&80.73\\
		&DAN~\cite{Long2015Learning}&46.44&53.56&58.60&42.68&65.66&65.34&55.38\\
		&DANN~\cite{ganin2016domain}&41.35&46.78&38.85&41.36&41.34&44.68&42.39\\
		&RTN~\cite{Long2016Unsupervised}&75.25&97.12&98.32&66.88&85.59&85.70&84.81\\
		&ADDA~\cite{Tzeng_2017_CVPR}&43.65&46.48&40.12&43.66&42.67&45.95&43.77\\
		&PADA~\cite{Cao_2018_ECCV} &86.54&\textbf{99.32}&\textbf{100.00}&82.17&\text92.69&\textbf{95.41}&92.69\\
		\cline{2-9}
		&DARL w/o Q-learning&84.07&96.61&\textbf{100.00}&85.35&79.75&78.81&87.43\\
		&DAL with pseudo labels&82.71&98.31&\textbf{100.00}&87.90&91.96&81.11&90.33\\
		&DARL &\textbf{90.17}&\textbf{99.32}&\textbf{100.00}&\textbf{90.45}&\textbf{93.42}&93.11&\textbf{94.41}\\
		\hline
	\end{tabular}
	\vspace{-2.5mm}
    \caption{Classification accuracies ($\%$) of partial domain adaptation tasks on Office-31.}
	\label{table:results on Office-31}
\end{table*}

\subsection{Discussion}
Most existing methods of partial domain adaptation utilize pseudo labels to weigh source instances in a straightforward manner~\cite{Cao_2018_ECCV,Cao_2018_CVPR,zhang2018importance}. 
In contrast, our DARL method applies the reinforcement learning paradigm to automatically learn policies for selecting source instances. 
The advantages of using reinforcement learning are as follows.
On one hand, reinforcement learning does not only make use of the prediction information but also explores in a wider space to find better solutions.
Since the agent is able to take actions of small Q-values with a certain probability, it has the ability of jumping out of local minima.
On the other hand, the selection strategy in DARL is a sequential decision process at the set level with the guidance of the accumulated rewards, it can be more accurate compared to selecting based on pseudo labels at the instance level.

\section{Experiments}
We compare our method with a number of baselines: AlexNet with bottleneck~\cite{Krizhevsky2012ImageNet}, ResNet with bottleneck~\cite{Kaiming_2016_CVPR}, Deep Adaptation Network (DAN)~\cite{Long2015Learning}, Domain-Adversarial Training of Neural Networks (DANN)~\cite{ganin2016domain},
Residual Transfer Network (RTN)~\cite{Long2016Unsupervised}, Adversarial
Discriminative Domain Adaptation (ADDA)~\cite{Tzeng_2017_CVPR}, Importance Weighted Adversarial Nets (IWAN)~\cite{zhang2018importance}, Selective Adversarial Networks (SAN)~\cite{Cao_2018_CVPR}, Partial Adversarial Domain Adaptation (PADA)~\cite{Cao_2018_ECCV}, where SAN, IWAN and PADA are proposed for partial domain adaptation.
\begin{table}
	\centering
	\scriptsize
	\vspace{-2.5mm}
	\begin{tabular}{|l|c|c|c|c|}
		\hline
		\multirow{3}{*}{Method} &\multicolumn{4}{c|}{Caltech-Office}\\
		\cline{2-5}
		& C256 $\to$ & C256$\to$& C256 $\to$ &\multirow{2}{*}{Avg}\\
		& W10&A10&D10&\\
		\hline
		AlexNet+bottleneck &62.37&78.39&65.61&68.79\\
		DAN~\cite{Long2015Learning} &42.37&70.75&47.04&53.39\\
		DANN~\cite{ganin2016domain} &54.57&72.86&57.96&61.80\\
		RTN~\cite{Long2016Unsupervised} &71.02&81.32&62.35&71.56\\
		ADDA~\cite{Tzeng_2017_CVPR} &73.66&78.35&74.80&75.60\\
		IWAN~\cite{zhang2018importance} &86.10&82.25&84.08&84.14\\
		SAN~\cite{Cao_2018_CVPR} &\textbf{88.33}&83.82&85.35&85.83\\
		\hline
		DARL w/o Q-learning &63.05&78.50&65.61&69.05\\
		DAL with pseudo labels &83.73&92.28&87.26&87.76\\
		DARL &88.14&\textbf{92.59}&\textbf{91.72}&\textbf{90.82}\\
		\hline
	\end{tabular}
	\caption{Classification accuracies ($\%$) of partial domain adaptation tasks on Office-31 and Caltech-Office (AlexNet as base network).}
	\label{table:results on Caltech-Office}
\end{table}

\subsection{Datasets}
We conduct extensive experiments on the following three benchmark datasets.

\noindent\textbf{Office-31}~\cite{Saenko2010Adapting} includes 31 classes of 4652 images, including three domains: Amazon, DSLR, and Webcam. The Amazon (A) contains 2817 images downloaded from online merchants (www.amazon.com). The DSLR (D) contains 498 high resolution images taken by a digital SLR camera. 
The Webcam (W) contains 795 low resolution images taken by a web camera. The three domains with total 31 classes (A31, D31 and W31) are used as source domains. The ten common classes of Office-31 and Caltech-256~\cite{GriffinGS2007Caltech} of Office-31 (A10, D10 and W10) are used as target domains. There are six transfer tasks: A31 $\to$ W10, D31 $\to$ W10, W31 $\to$ D10, A31 $\to$ D10, D31 $\to$ A10 and W31 $\to$ A10.

\noindent\textbf{Office+Caltech-10}~\cite{Gong2012Geodesic} has four domains: Amazon (A), DSLR (D), Webcam (W) and Caltech (C), including ten common classes of Office-31 and Caltech-256. The four domains (A10, D10, W10 and C10) are used as source domains. Following the setting of~\cite{zhang2018importance}, the first five classes (``back pack", ``bike", ``calculator", ``headphones" and ``keyboard") of the four domains are used as the target domains (A5, D5, W5 and C5). There are 12 transfer tasks: C10 $\to$ A5, C10 $\to$ W5, C10 $\to$ D5, A10 $\to$ C5, A10 $\to$ W5, A10 $\to$ D5, W10 $\to$ C5, W10 $\to$ A5, W10 $\to$ D5, D10 $\to$ C5, D10 $\to$ A5, D10 $\to$ W5.

\noindent\textbf{Caltech-Office} is constructed with Caltech-256 and Office-31. Caltech-256 consists of 30607 images in 256 categories, collecting from Google and PicSearch. The Caltech-256 is used as source domain, denoted as C256, and the ten shared classes of Office-31 and Caltech-256 (A10, D10 and W10) are used as target domains. There are three transfer tasks: C256 $\to$ W10, C256 $\to$ A10 and C256 $\to$ D10.

\subsection{Implementation Details}
\label{setup}

Following the setting of SAN and IWAN, we fine-tune from the AlexNet model pre-trained on the ImageNet dataset.  Concretely, the feature extractor $F$ is obtained by removing the \emph{fc8} layer of AlexNet and adding a bottleneck layer with 256 units on \emph{fc7}. We fine-tune the \emph{conv5}, \emph{fc6} and \emph{fc7} layers of $F$, and train the bottleneck layer of $F$ and the classifier $C$. The bottleneck layer of $F$ and the classifier $C$ are trained from scratch, whose learning rate is set to be 10 times of the other layers~\cite{Cao_2018_CVPR}.
The discriminator $D$ is built with three \emph{fc} layers (1024 $\to$ 1024 $\to$ category number+1). 
The deep Q-learning network has four \emph{fc} layers (1024 $\to$ 512 $\to$ 256 $\to$ action number).
Following the setting of PADA, we fine-tune from the ResNet-50 model pre-trained on the ImageNet dataset.
The feature extractor $F$ is obtained by removing the \emph{fc} layer of ResNet and adding a bottleneck layer with 256 units on \emph{res5c}. The training strategy is the same as AlexNet.

We apply the $\epsilon$-greedy strategy~\cite{Mnih2015Human} and the experience replay strategy~\cite{Lin1992Self} to the deep Q-learning.
The AdamOptimizer is used to optimize the whole network. 
The learning rates of deep Q-learning network and domain adversarial learning network are both set to 0.0001 with 0.9 and 0.5 as the momentum, respectively. 
The discount factor $\gamma$ is set to 0.9. 
During the exploration stage of the deep Q-learning, the exploration rate is decayed from 1 to 0. 
The threshold $\tau$ is set to 0.3 and 0.1 for AlexNet and ResNet as base network, respectively, detailed analyzed in Section ~\ref{section:Analysis}.

\subsection{Results}
The classification accuracies of different methods on the Office+Caltech-10, Office-31 and Caltech-Office datasets are reported in Table \ref{table:results on Office+Caltech-10}, Table \ref{table:results on Office-31} and Table \ref{table:results on Caltech-Office}. For all the  compared methods, we directly use the reported results in their original papers to make the comparison fair. 
From the results, we have the following observations:
\begin{itemize}
    \item{DARL outperforms all the compared methods on most transfer tasks, clearly demonstrating the benefit of reinforcement learning on selecting right source instances for partial domain adaptation.}
    \item{DARL substantially promotes the classification accuracy especially on the difficult Catlech-Office dataset, where there is a large gap between the label spaces of different domains. It outperforms SAN and IWAN with the gains of 4.65$\%$ and 6.34$\%$ on average,respectively, which verifies that DARL is excellent in handling more challenging partial domain adaptation.}
    \item{For both Alexnet and ResNet-50 base models, DARL consistently achieves better results than other methods, which validates the superior generalization ability of DARL.}
\end{itemize}

\subsection{Analysis}
\label{section:Analysis}
\begin{figure}
		\centering
		\subfigure[Accuracy vs threshold]{ \includegraphics[width=0.225\textwidth]{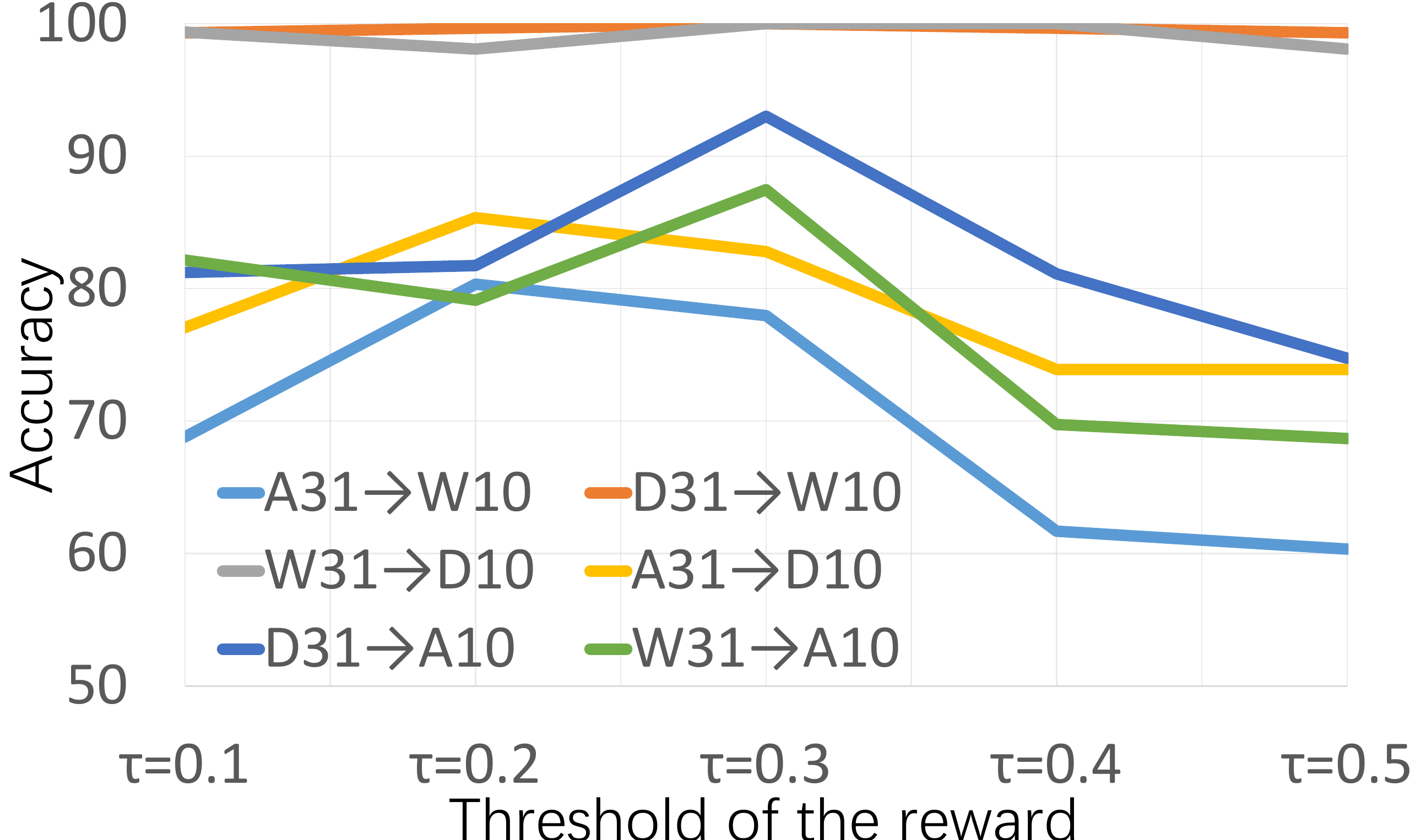}
			\label{fig:threshold}}
		\subfigure[Test error vs iteration]{			\includegraphics[width=0.23\textwidth]{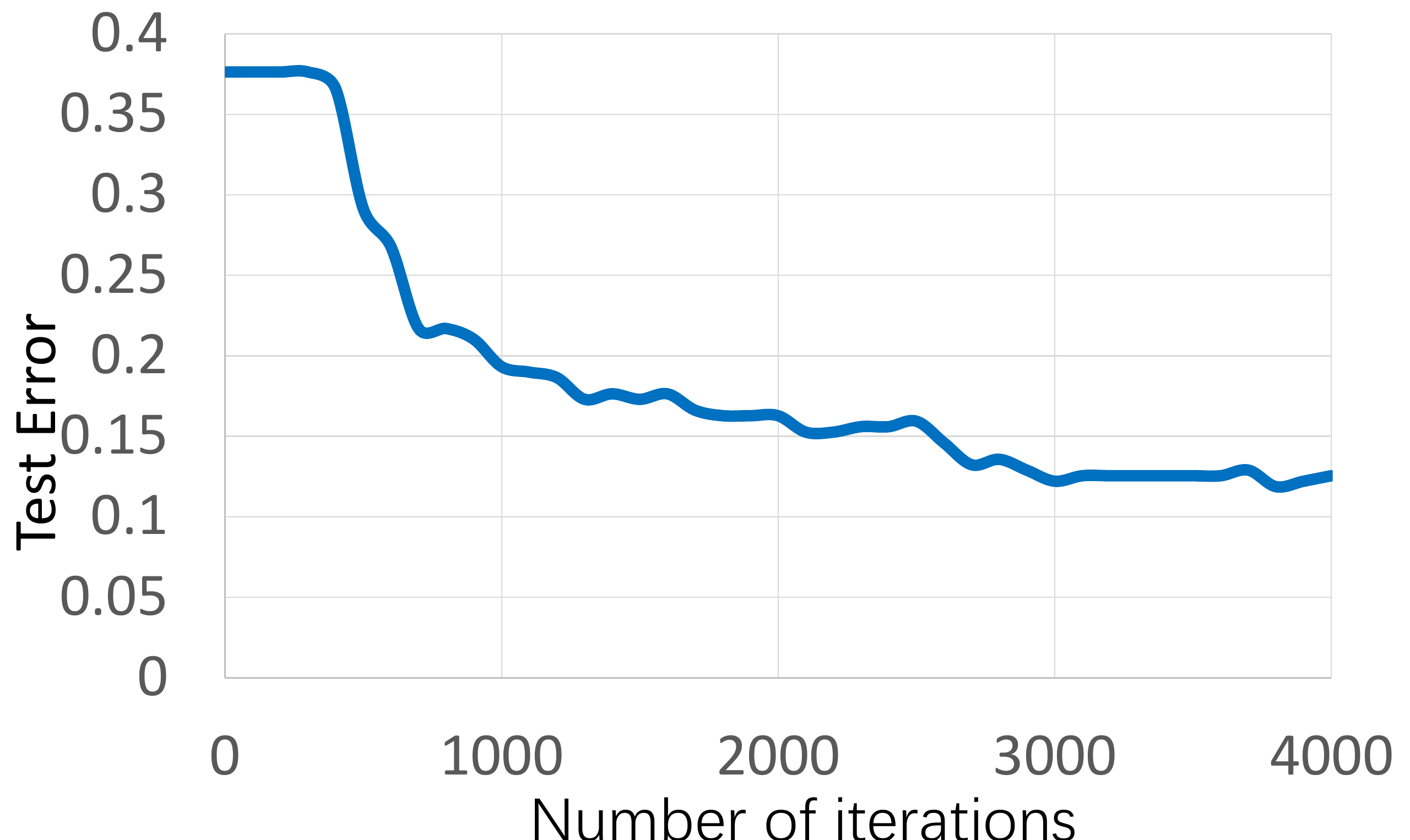}
			\label{fig:test-error}}
		\caption{Empirical analysis of DARL. (a) Performance of different threshold. (b) Test error on target domain iterations }
		\label{fig:metricvisualization}
\end{figure}

\noindent\textbf{Parameter analysis.}
The threshold of reward function $\tau$ is an import factor in DARL. More experiments are conducted with different values of $\tau$ on the Office-31 dataset with AlexNet as base network. From Figure \ref{fig:threshold}, it is interesting to observe that the accuracies of all the tasks first increase and then decrease with the increasing threshold $\tau$. Specifically, when $\tau$ is small, the accuracies are lower since some source instances in the outlier classes can not be filtered out. 
When $\tau$ is large, the accuracies are also lower since some source instances in the shared classes are filtered out. Thus, we set $\tau = 0.3$ for the best results on most transfer task with AlexNet as base network. With the same strategy, we set $\tau = 0.1$ with ResNet as base network.

\noindent\textbf{Ablation study.}
To go deeper with each component of DARL, we compare our method with two variations: without deep Q-learning (DARL w/o Q-learning), replace deep Q-learning with pseudo labels strategy (Domain Adversarial learning with pseudo labels, \emph{i.e.,} DAL with pseudo labels). 
In the method of DAL with pseudo labels, source instances are selected only by the relevance metric. Concretely, the source instances with $\varphi(x)> \tau$ are selected as training data for domain adversarial learning, where we set the same threshold $\tau$ as DARL.
From the results shown in Table \ref{table:results on Office+Caltech-10}, \ref{table:results on Office-31}, and \ref{table:results on Caltech-Office}, DARL outperforms the method of DARL w/o Q-learning, which clearly validates the benefit of deep Q-learning on selecting the right source instances for positive transfer.
When removing the deep Q-learning from DARL, the classification accuracies will substantially degrade.
DARL also works better than the method of DAL with pseudo labels on all the transfer tasks, demonstrating that reinforcement learning is more powerful than selecting with the pseudo label directly.

\noindent\textbf{Convergence performance.}
We study the test error on the C256 $\to$ W10 task to evaluate the convergence performance of DARL. In Figure \ref{fig:test-error}, it can be observed that DARL can gradually converge to a low test error.

\section{Conclusion}
In this paper, we have proposed a reinforcement learning framework coupled with adversarial learning for partial domain adaptation. The deep Q-learning component can select source instances in the shared classes for avoiding negative transfer. The domain adversarial learning component can reduce the domain shift and provide effective rewards to the agent for promoting positive transfer. The two components are jointly learned by an iterative optimization to make them promote each other. Extensive experiments on various benchmarks have demonstrated the effectiveness of the proposed method.

\bibliographystyle{named}
\bibliography{ijcai19}

\begin{thebibliography}{}

\bibitem[\protect\citeauthoryear{Cao \bgroup \em et al.\egroup
  }{2018a}]{Cao_2018_CVPR}
Zhangjie Cao, Mingsheng Long, Jianmin Wang, and Michael~I. Jordan.
\newblock Partial transfer learning with selective adversarial networks.
\newblock In {\em Proceedings of the IEEE Conference on Computer Vision and
  Pattern Recognition (CVPR)}, pages 2724--2732, June 2018.

\bibitem[\protect\citeauthoryear{Cao \bgroup \em et al.\egroup
  }{2018b}]{Cao_2018_ECCV}
Zhangjie Cao, Lijia Ma, Mingsheng Long, and Jianmin Wang.
\newblock Partial adversarial domain adaptation.
\newblock In {\em Proceedings of the European Conference on Computer Vision
  (ECCV)}, September 2018.

\bibitem[\protect\citeauthoryear{Dong and Xing}{2018}]{dong2018domain}
Nanqing Dong and Eric~P. Xing.
\newblock Domain adaption in one-shot learning.
\newblock In {\em The European Conference on Machine Learning and Principles
  and Practice of Knowledge Discovery in Databases (ECML-PKDD)}, pages
  573--588, 2018.

\bibitem[\protect\citeauthoryear{Ganin \bgroup \em et al.\egroup
  }{2016}]{ganin2016domain}
Yaroslav Ganin, Evgeniya Ustinova, Hana Ajakan, Pascal Germain, Hugo
  Larochelle, Fran{\c{c}}ois Laviolette, Mario Marchand, and Victor Lempitsky.
\newblock Domain-adversarial training of neural networks.
\newblock {\em Journal of Machine Learning Research (JMLR)}, 17(1):2096--2030,
  2016.

\bibitem[\protect\citeauthoryear{Gong \bgroup \em et al.\egroup
  }{2012}]{Gong2012Geodesic}
Boqing Gong, Yuan Shi, Fei Sha, and Kristen Grauman.
\newblock Geodesic flow kernel for unsupervised domain adaptation.
\newblock In {\em Proceedings of the Computer Vision and Pattern Recognition
  (CVPR)}, pages 2066--2073, 2012.

\bibitem[\protect\citeauthoryear{GriffinGS \bgroup \em et al.\egroup
  }{2007}]{GriffinGS2007Caltech}
GriffinGS, HolubAD, and PeronaP.
\newblock Caltech-256 object category dataset.
\newblock {\em California Institute of Technology}, 2007.

\bibitem[\protect\citeauthoryear{He \bgroup \em et al.\egroup
  }{2016}]{Kaiming_2016_CVPR}
Kaiming He, Xiangyu Zhang, ShaoqingRen, and Jian Sun.
\newblock Deep residual learning for image recognition.
\newblock In {\em The IEEE Conference on Computer Vision and Pattern
  Recognition (CVPR)}, pages 770--778, June 2016.

\bibitem[\protect\citeauthoryear{Hu \bgroup \em et al.\egroup
  }{2018}]{hu2018duplex}
Lanqing Hu, Meina Kan, Shiguang Shan, and Xilin Chen.
\newblock Duplex generative adversarial network for unsupervised domain
  adaptation.
\newblock In {\em Proceedings of the IEEE Conference on Computer Vision and
  Pattern Recognition (CVPR)}, pages 1498--1507, 2018.

\bibitem[\protect\citeauthoryear{Huang \bgroup \em et al.\egroup
  }{2018}]{Huang2017A}
Jingjia Huang, Nannan Li, Tao Zhang, and Ge~Li.
\newblock A self-adaptive proposal model for temporal action detection based on
  reinforcement learning.
\newblock In {\em Proceedings of the Association for the Advancement of
  Artificial Intelligence (AAAI)}, pages 6951--6958, 2018.

\bibitem[\protect\citeauthoryear{Krizhevsky \bgroup \em et al.\egroup
  }{2012}]{Krizhevsky2012ImageNet}
Alex Krizhevsky, Ilya Sutskever, and Geoffrey~E Hinton.
\newblock Imagenet classification with deep convolutional neural networks.
\newblock In {\em Advances in neural information processing systems (NIPS)},
  pages 1097--1105, 2012.

\bibitem[\protect\citeauthoryear{Lin}{1992}]{Lin1992Self}
Long~Ji Lin.
\newblock Self-improving reactive agents based on reinforcement learning,
  planning and teaching.
\newblock {\em Machine Learning}, 8(3-4):293--321, 1992.

\bibitem[\protect\citeauthoryear{Long \bgroup \em et al.\egroup
  }{2015}]{Long2015Learning}
Mingsheng Long, Yue Cao, Jianmin Wang, and Michael~I. Jordan.
\newblock Learning transferable features with deep adaptation networks.
\newblock In {\em Proceedings of the International Conference on Machine
  Learning (ICML)}, pages 97--105, 2015.

\bibitem[\protect\citeauthoryear{Long \bgroup \em et al.\egroup
  }{2016}]{Long2016Unsupervised}
Mingsheng Long, Han Zhu, Jianmin Wang, and Michael~I Jordan.
\newblock Unsupervised domain adaptation with residual transfer networks.
\newblock In {\em Advances in Neural Information Processing Systems (NIPS)},
  pages 136--144, 2016.

\bibitem[\protect\citeauthoryear{Mnih \bgroup \em et al.\egroup
  }{2015}]{Mnih2015Human}
Volodymyr Mnih, Koray Kavukcuoglu, David Silver, Andrei~A Rusu, Joel Veness,
  Marc~G Bellemare, Alex Graves, Martin Riedmiller, Andreas~K Fidjeland, Georg
  Ostrovski, et~al.
\newblock Human-level control through deep reinforcement learning.
\newblock {\em Nature}, 518(7540):529, 2015.

\bibitem[\protect\citeauthoryear{Pirinen and
  Sminchisescu}{2018}]{Pirinen_2018_CVPR}
Aleksis Pirinen and Cristian Sminchisescu.
\newblock Deep reinforcement learning of region proposal networks for object
  detection.
\newblock In {\em The IEEE Conference on Computer Vision and Pattern
  Recognition (CVPR)}, June 2018.

\bibitem[\protect\citeauthoryear{Ren \bgroup \em et al.\egroup
  }{2018a}]{Ren_2018_ECCV_Collaborative}
Liangliang Ren, Jiwen Lu, Zifeng Wang, Qi~Tian, and Jie Zhou.
\newblock Collaborative deep reinforcement learning for multi-object tracking.
\newblock In {\em The European Conference on Computer Vision (ECCV)}, September
  2018.

\bibitem[\protect\citeauthoryear{Ren \bgroup \em et al.\egroup
  }{2018b}]{Ren_2018_ECCV}
Liangliang Ren, Xin Yuan, Jiwen Lu, Ming Yang, and Jie Zhou.
\newblock Deep reinforcement learning with iterative shift for visual tracking.
\newblock In {\em Proceedings of the European Conference on Computer Vision
  (ECCV)}, September 2018.

\bibitem[\protect\citeauthoryear{Saenko \bgroup \em et al.\egroup
  }{2010}]{Saenko2010Adapting}
Kate Saenko, Brian Kulis, Mario Fritz, and Trevor Darrell.
\newblock Adapting visual category models to new domains.
\newblock In {\em Proceedings of the European Conference on Computer Vision
  (ECCV)}, pages 213--226, 2010.

\bibitem[\protect\citeauthoryear{Sankaranarayanan \bgroup \em et al.\egroup
  }{2018}]{sankaranarayanan2018generate}
Swami Sankaranarayanan, Yogesh Balaji, Carlos~D Castillo, and Rama Chellappa.
\newblock Generate to adapt: Aligning domains using generative adversarial
  networks.
\newblock In {\em Proceedings of the IEEE Conference on Computer Vision and
  Pattern Recognition (CVPR)}, pages 8503--8512, 2018.

\bibitem[\protect\citeauthoryear{Shen \bgroup \em et al.\egroup
  }{2018}]{Shen2018Wasserstein}
Jian Shen, Yanru Qu, Weinan Zhang, and Yong Yu.
\newblock Wasserstein distance guided representation learning for domain
  adaptation.
\newblock In {\em Proceedings of the Association for the Advancement of
  Artificial Intelligence (AAAI)}, pages 4058--4065, 2018.

\bibitem[\protect\citeauthoryear{Sutton and
  Barto}{1998}]{Sutton1998Reinforcement}
Richard~S Sutton and Andrew~G Barto.
\newblock Reinforcement learning: An introduction.
\newblock {\em IEEE Transactions on Neural Networks and Learning Systems
  (TNNLS)}, 9(5):1054--1054, 1998.

\bibitem[\protect\citeauthoryear{Tzeng \bgroup \em et al.\egroup
  }{2017}]{Tzeng_2017_CVPR}
Eric Tzeng, Judy Hoffman, Kate Saenko, and Trevor Darrell.
\newblock Adversarial discriminative domain adaptation.
\newblock In {\em The IEEE Conference on Computer Vision and Pattern
  Recognition (CVPR)}, pages 7167--7176, July 2017.

\bibitem[\protect\citeauthoryear{Wang \bgroup \em et al.\egroup
  }{2018}]{Wang_2018_CVPR}
Xin Wang, Wenhu Chen, Jiawei Wu, Yuan-Fang Wang, and William Yang~Wang.
\newblock Video captioning via hierarchical reinforcement learning.
\newblock In {\em The IEEE Conference on Computer Vision and Pattern
  Recognition (CVPR)}, June 2018.

\bibitem[\protect\citeauthoryear{Yeung \bgroup \em et al.\egroup
  }{2017}]{Yeung_2017_CVPR}
Serena Yeung, Vignesh Ramanathan, Olga Russakovsky, Liyue Shen, Greg Mori, and
  Li~Fei-Fei.
\newblock Learning to learn from noisy web videos.
\newblock In {\em Proceedings of the IEEE Conference on Computer Vision and
  Pattern Recognition (CVPR)}, pages 5154--5162, July 2017.

\bibitem[\protect\citeauthoryear{Yun \bgroup \em et al.\egroup
  }{2017}]{Yun2017Action}
Sangdoo Yun, Jongwon Choi, Youngjoon Yoo, Kimin Yun, and Jin Young~Choi.
\newblock Action-decision networks for visual tracking with deep reinforcement
  learning.
\newblock In {\em Proceedings of the IEEE Conference on Computer Vision and
  Pattern Recognition (CVPR)}, pages 1349--1358, July 2017.

\bibitem[\protect\citeauthoryear{Zhang \bgroup \em et al.\egroup
  }{2018}]{zhang2018importance}
Jing Zhang, Zewei Ding, Wanqing Li, and Philip Ogunbona.
\newblock Importance weighted adversarial nets for partial domain adaptation.
\newblock In {\em Proceedings of the IEEE Conference on Computer Vision and
  Pattern Recognition (CVPR)}, pages 8156--8164, 2018.

\end{thebibliography}
\end{document}